\documentclass[preprint,12pt]{elsarticle}

\usepackage{makecell}
\usepackage{amssymb}
\usepackage{amsmath}
\usepackage{hyperref} 
\usepackage{threeparttable} 
\usepackage{stackengine}
\usepackage{booktabs} 
\usepackage{url}

\journal{Nuclear Physics B}

\begin{document}

\begin{frontmatter}



\title{Robust Cross-Domain Generalization Using Unlabeled Target Data with Source-Domain Supervision}


\author{Yuyue Zhou, Shrimanti Ghosh, Michael (Kai Yue) Xie, Justin JY Kim, Jessica Knight, Steel McDonald, Vincent Man, Jacob L. Jaremko, Abhilash Hareendranathan} 

\affiliation{organization={Department of Radiology and Diagnostic Imaging, University of Alberta},
            addressline={8303 112 St NW}, 
            city={Edmonton},
            postcode={T6G 2T4}, 
            state={AB},
            country={Canada}}

\begin{abstract}
It is often desirable to generalize medical imaging AI models trained with dense annotations to data acquired from different ultrasound scanners or clinical sites; however, retraining these models with new annotations is often difficult and costly. We examine this challenge in pediatric wrist fracture assessment using point-of-care ultrasound (POCUS), where fractures are common and can be effectively triaged via ultrasound. AI has shown radiologist-level performance for fracture detection, often aided by high-quality bony structure segmentation. However, due to significant domain shifts, models perform poorly on data from other centers or probes, and obtaining segmentation labels across devices is impractical due to manual annotation effort and data privacy concerns. To address this, we propose a target-informed self-supervised pretraining and model-ensemble strategy. Specifically, our approach combines masked image modeling (MIM) and contrastive learning to learn target-domain structural representations without labels, and introduces a confidence-aware infusion head to adaptively integrate predictions. The source dataset, collected with a Philips Lumify probe, contained dense labels, while the target dataset, acquired with a TeleMED portable probe, was unlabeled. The datasets were kept strictly separate throughout the entire process. Our method used labeled source data for supervised training and leveraged target-domain pretraining to improve generalization. On 318 images from 62 pediatric POCUS videos, this approach significantly improved cross-device performance, achieving over 6\% Dice improvement on the target domain versus the baseline. These results demonstrate a label-efficient and privacy-preserving approach for cross-device–robust ultrasound AI, offering a framework that can be extended to multi-center studies or federated learning setups.

\end{abstract}


\end{frontmatter}


\noindent\textbf{GitHub Repository:} 
\href{https://github.com/yuyue2uofa/CrossDomainPOCUS}{https://github.com/yuyue2uofa/CrossDomainPOCUS}

\section{Background}

Artificial intelligence (AI) has become increasingly influential in medical imaging, demonstrating strong performance across tasks such as classification \cite{ghosh2024ultrasound}, detection \cite{ghosh2024automated}, and segmentation \cite{zhou2022wrist}. Most of these models are trained via supervised learning on single-center datasets acquired from uniform imaging devices. Medical images acquired from different scanners or clinical sites often exhibit substantial variations in appearance due to differences in hardware, acquisition protocols, and operator settings. As a result, the performance of most AI models reduces significantly when tested on these new datasets. This phenomenon is commonly referred to as domain shift. Domain shift refers to the degradation in model performance when a model trained on one dataset is applied to another with a different data distribution \cite{pan2009survey}. Prior studies have demonstrated that models trained on a single dataset often generalize poorly to data from different sources, primarily due to dataset-specific biases and distributional discrepancies \cite{torralba2011unbiased, david2010impossibility}. 

In practice, these models require retraining on new annotations, which involves significant costs and extensive expert involvement. Hence it is desirable to have AI models that generalize to new medical imaging datasets acquired from different scanners or clinical sites without retraining with new  annotations. In medical imaging, domain shift is particularly pervasive as a result of variations in imaging devices, acquisition protocols, and patient populations, posing a significant challenge for developing robust and generalizable models \cite{guan2021domain}. This issue is particularly pronounced in ultrasound imaging, where inter-scanner and inter-center variability markedly degrades model generalizability, motivating the development of domain-bridging and adaptation methods \cite{huang2024standardization, wu2024improving}.

In this work, we study this problem in the context of pediatric wrist fracture assessment.

\subsection{Wrist Fracture}

Wrist fractures are the most common fractures among children, accounting for nearly one-third of all pediatric fractures \cite{brudvik2003childhood, hedstrom2010epidemiology}. They are typically caused by accidental fall on the outstretched hand (FOOSH) injuries, resulting in symptoms such as swelling, pain, tenderness, and reduced joint mobility. Management depends heavily on fracture severity and location: mild fractures may be treated with immobilization in a cast or splint, whereas more severe or displaced fractures often require surgical intervention \cite{zhu2024non}. Consequently, timely and accurate diagnosis is essential to ensure appropriate treatment planning and to prevent long-term complications.

Currently, the clinical gold standard for diagnosing wrist fractures is radiography \cite{backer2020systematic}. Despite its effectiveness, radiography requires large, non-portable equipment that must be operated in rooms with proper radiation protection. This lack of portability limits its use outside specialized hospital or tertiary care settings. As a result, patients with suspected wrist injuries must seek evaluation in hospital emergency departments or imaging centers, which create barriers to accessibility for families living far away from such medical institutions. These limitations also contribute to significant delays in care: in Canada, the average wait time in the emergency departments exceeds 8 hours, and nearly half of these patients are ultimately found not to have a fracture \cite{CIHI2024, slaar2016clinical}. Together, these factors highlight the need for an accessible and efficient frontline alternative.

Point-of-care ultrasound (POCUS) has emerged as a promising portable alternative for wrist fracture assessment. Studies have shown that radiologists can detect wrist fractures using ultrasound or POCUS with accuracy comparable to radiographs \cite{knight20232d, zhou2025wrist}. Compared with traditional X-ray imaging, POCUS is radiation-free, lightweight, cost-effective, easy to transport, and capable of real-time visualization. Importantly, it can be operated by personnel with basic training, including triage nurses or even non-medical users, making it well-suited for high-demand or resource-limited settings. With accurate interpretation, POCUS could help ensure that only patients with confirmed fractures are referred to hospitals for further care, reducing wait times, lowering healthcare costs, and alleviating systemic burden.

\begin{figure}[t]
    \centering  
    \includegraphics[width=\linewidth]{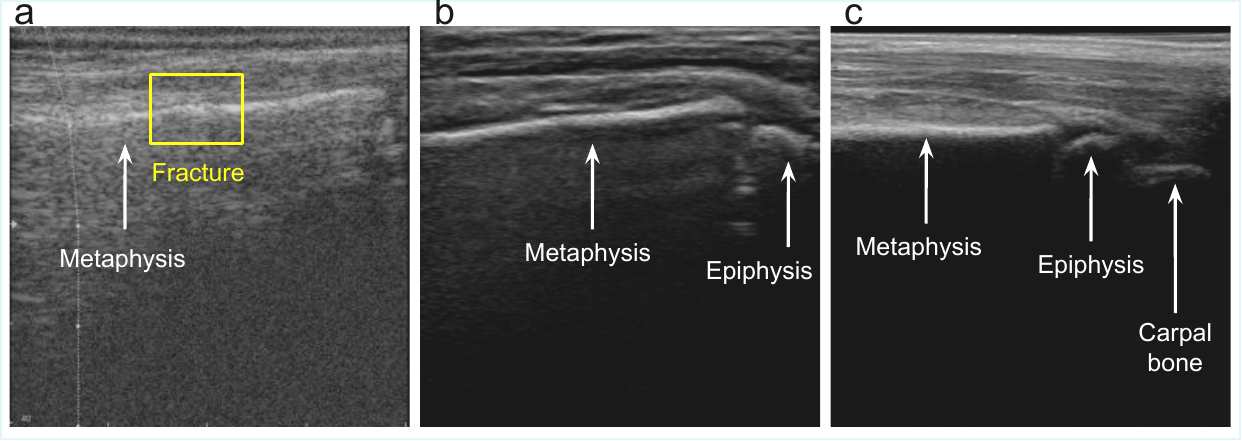}  
    \caption{Wrist ultrasound images acquired using 3 different probes. a: POCUS TeleMED MicrUs Pro-L40S ultrasound probe; b: POCUS Philips Lumify L5–12 MHz  ultrasound probe; c: conventional cart-based ultrasound system, Philips iU22 machine with a 13 MHz VL13-5 probe. The arrow denotes the relevant bone structures, and the fracture area is highlighted in yellow. Compared with the conventional ultrasound system, POCUS images exhibit lower image quality, particularly those from the TeleMED probe. In addition, there remains a noticeable domain shift among POCUS images acquired with different probes.} 
    \label{fig1} 
\end{figure}

Despite these advantages, interpreting POCUS images remains challenging, as accurate diagnosis still relies heavily on trained radiologists. Non-medical operators, although able to acquire images, often cannot reliably distinguish between bone surfaces, cortical disruptions, and surrounding soft-tissue structures \cite{chartier2017use}. Furthermore, compared with traditional cart-based or high-resolution ultrasound systems (e.g., Philips iU22 machine with a 13 MHz VL13-5 probe), some POCUS devices, such as the TeleMED MicrUs Pro-L40S probe, typically produce images with blurrier anatomical boundaries, lower contrast, and more noise or artifacts (Figure \ref{fig1}). These quality limitations make musculoskeletal structures, particularly bone edges, more difficult to interpret, even for experienced clinicians. Addressing these challenges is essential for broader clinical adoption of ultrasound in musculoskeletal imaging, especially in settings where expert interpretation is not readily available. Segmenting the bony regions in the image like metaphysis and epiphysis is often the first step in an image analysis pipeline. Various image segmentation approaches have been proposed to  automate this step and reduce the dependence on experts for interpretation.

Building upon this pipeline, in a real-world deployment scenario where a new unlabeled POCUS device is introduced in a clinic, our framework can be applied without manual annotation to enable rapid adaptation, while providing timely bone segmentation to support fracture assessment by delineating bone structures.

\subsection{Artificial Intelligence Segmentation and Our Scenario}

The advancement of AI has introduced new opportunities for automating medical image interpretation, including ultrasound imaging, where expert knowledge is often required for accurate diagnosis. AI-driven approaches have the potential to reduce reliance on specialized expertise while improving diagnostic efficiency and accuracy. In particular, accurate segmentation is critical for disease detection, severity assessment, and subsequent treatment planning \cite{pinto2023artificial}. In the context of musculoskeletal ultrasound, numerous AI-based segmentation methods have demonstrated the feasibility of accurately delineating anatomical structures such as nerves \cite{di2022deep, smerilli2022development} cartilage \cite{du2022automatic}, and muscles \cite{marzola2021deep}, and facilitated disease diagnosis. Building upon these advances, prior studies have shown that precise segmentation of bony or related structures can further enhance AI performance in musculoskeletal ultrasound, enabling more reliable identification of bone-related abnormalities such as shoulder injuries \cite{ghosh2024ultrasound}, wrist and elbow fracture \cite{zhou2025wrist}, hip dysplasia \cite{lee2021accuracy}, arthritis \cite{hemalatha2019automatic}, scoliosis \cite{lyu2021dual} as well as supporting bone-related treatment planning in applications such as computer-assisted orthopedic surgery \cite{alsinan2019automatic, luan2020efficient, wang2020robust}.

AI-based segmentation requires large amounts of annotated data. For ultrasound, however, a single video may contain dozens to hundreds of frames, each of which must be annotated by a trained medical professional. This process is extremely time-consuming and costly. Furthermore, distinct appearance characteristics across different ultrasound probes can impair model generalizability. This necessitates robust AI solutions capable of maintaining performance across diverse datasets with minimal additional annotations.

In this work, we consider a practical musculoskeletal ultrasound scenario involving data collected from different probes at different time points, where labeled data are available in one domain but absent in another. In our experimental setup, we simulate a scenario in which data cannot be shared due to privacy concerns by treating the two datasets as isolated during training. Consequently, inspired by federated learning, we adopt an approach in which model weights are transferred instead of data. Our main goal is to train a segmentation model using only the annotations from the source domain while ensuring strong performance on the domain-shifted, unlabeled target domain.  In addition to this primary objective, we also ensure that the model remains robust on the source domain test set.

Unlike conventional approaches for handling domain shift in medical image analysis, including continuous learning, transfer learning, domain adaptation, and multi-source training, we adopt a cross-domain self-supervised learning (SSL) strategy, where unlabeled target-domain data are leveraged for representation pretraining, followed by supervised fine-tuning on the source domain to improve robustness under domain shift.

This strategy addresses the challenges posed by the labor-intensive nature of ultrasound labeling, enables robustness under domain shift, and respects privacy constraints. Through this approach, we aim to bridge the gap between state-of-the-art AI segmentation techniques and practical, real-world musculoskeletal ultrasound applications.

\section{Related Work}

Building on the domain shift challenges discussed earlier, this section reviews prior studies that aim to improve cross-domain generalization in medical imaging.

\subsection{Existing Cross-Domain Approaches in Medical Imaging Segmentation}

Traditional approaches have applied transfer learning in medical image segmentation using pre-trained models from related domains, typically fine-tuning with a small number of target-domain labels \cite{zoetmulder2022domain, ghafoorian2017transfer}. Meta-learning has also been explored for medical image segmentation \cite{zhang2021domain}. However, these strategies are ineffective when the target domain has no annotations. 

Recent studies have explored diffusion-based and CycleGAN-based approaches for annotation-free cross-domain medical image segmentation. Yang et al. employ diffusion models to learn domain-independent conditional distributions, enabling robust cross-domain medical image segmentation \cite{yang2025domain}. Zeng et al. leverage CycleGAN to enforce both intra- and cross-modality semantic consistency, thereby improving segmentation performance in CT and MRI domains \cite{zeng2021semantic}. Wang et al. combine cycle-consistent image translation with semantic consistency constraints to align source and target domains without requiring target annotations \cite{wang2022cycmis}. Jiang et al. further reduce domain discrepancy in CT segmentation through image-level style translation and feature-level feature alignment \cite{jiang2023medical}. However, these methods share several key limitations, including high model complexity, prolonged training time, and substantial computational cost, which lead to inefficient inference and hinder their practical deployment in real-world clinical settings.

Notably, foundation models such as SAM \cite{kirillov2023segment} achieve strong results on natural images but perform poorly in medical imaging under cross-domain conditions \cite{shi2023generalist}. Even domain-specific foundation models like MedSAM \cite{ma2024segment} still exhibit suboptimal performance when applied to unseen domains \cite{noh2025narrative}. Recent studies have shown that this limitation extends beyond medical imaging, where foundation models are not readily transferable under domain shift and require additional adaptation mechanisms, such as LoRA, for effective deployment in target domains \cite{ning20252}.

These observations highlight the critical need for methods that can achieve cross-domain robustness, particularly in settings where the target domain lacks labels. This motivates new approaches such as target-domain self-supervised pretraining, which we adopt in this work to extract structural information from the unlabeled target data while training primarily on source-domain annotations.

\subsection{SSL and Its Applications in Cross-Domain Challenges}

SSL is an emerging technique that enables models to leverage unlabeled datasets \cite{balestriero2023cookbook}. It typically consists of two stages. In the first stage, the model is pretrained on a fully unlabeled dataset using a pretext task, allowing it to learn meaningful image features and representations. In the second stage, the pretrained weights are used to initialize the model, either fully or partially, which is then fine-tuned on a small subset of labeled data for downstream tasks such as segmentation, classification, or depth estimation \cite{wolf2023self}. SSL methods can be grouped into two major families: contrastive approaches and generative approaches \cite{balestriero2023cookbook}. Contrastive SSL learns representations by comparing positive and negative pairs \cite{chen2020simple, he2020momentum, grill2020bootstrap, chen2021exploring}, whereas generative SSL encompasses tasks that reconstruct or predict missing, corrupted, or transformed aspects of the input, ranging from classical autoencoding \cite{hinton2006reducing} and colorization \cite{larsson2017colorization} to more recent masked-image-modeling (MIM) techniques \cite{he2022masked, xie2022simmim}.

SSL has been used in medical imaging to address the cross-domain challenges. Zheng et al. \cite{zheng2021hierarchical} aggregated multiple datasets from different domains to pretrain and subsequently fine‑tune their model. Their results demonstrated that multi-domain datasets aggregation combined with SSL substantially improves segmentation performance. However, both the pretraining and fine‑tuning were performed on subsets of the aggregated datasets, and the study did not evaluate scenarios in which an entire domain has no available labels. Bundele et al. \cite{bundele2024evaluating} evaluated contrastive SSL under in-domain, cross-domain generalization, and out-of-domain robustness settings, showing consistent performance improvements from self-supervised pretraining. However, these approaches typically relied on target-domain labels during fine-tuning or did not explicitly investigate whether unlabeled target-domain data alone can be leveraged to improve cross-domain performance. He et al. \cite{he2021autoencoder} proposed a self-supervised test-time adaptation framework that introduces learnable adaptors optimized per test sample using autoencoder reconstruction loss to align target-domain features with the source-domain feature distribution, without updating the task model parameters. Ali et al. \cite{ali2025multimodal} proposed a multimodal SSL framework that jointly learns MRI and PET feature representations using cross-modal alignment, longitudinal consistency, and site-invariance objectives, demonstrating strong cross-cohort generalization and robustness to domain differences across multiple Alzheimer’s datasets. However, their approach relied on aggregating datasets from multiple cohorts for pretraining and fine-tuning, which is not always feasible in settings with strict data-sharing restrictions. Anton et al. \cite{anton2022well} systematically evaluated SSL’s ability on medical imaging datasets. They showed that SSL models, when both pretrained and fine-tuned on the same domain, excel on in-domain tasks but suffer notable performance loss when transferred to other datasets. 

In summary, existing SSL approaches in medical imaging cross-domain tasks largely rely on either target-domain labels during fine-tuning or on aggregating multiple datasets during pretraining to improve cross-domain generalization. Such strategies do not address the two challenges we focus on in this work: (1) the target domain is completely unlabeled during training, and (2) cross-domain datasets remain fully separate, with no aggregation permitted due to data governance or privacy constraints. This gap motivates the present study, in which we leverage multiple SSL methods and model-ensemble strategies in a privacy-preserving, cross-site setting with unlabeled target-domain data and substantial device-induced domain shift, achieving robust performance on both the source and target domains. 

\subsection{Our work}

In this study, we propose a novel SSL framework that integrates MIM and contrastive learning to enhance cross-domain performance in medical imaging tasks. To the best of our knowledge, this is the first work to explore heterogeneous self-supervised knowledge fusion, combining generative and contrastive SSL paradigms, within a cross-domain pretraining setting. Our main contributions are summarized as follows:

(1) A novel heterogeneous SSL knowledge-fusion architecture for cross-domain medical imaging.
We design a new SSL framework that jointly leverages two fundamentally different SSL strategies: a MIM-based TransUNet (MIM-TransUNet) and a new contrastive-learning TransUNet (Contrastive-TransUNet). The MIM-TransUNet adapts the principles of SimMIM to the TransUNet architecture, enabling masked image modeling in a U-shaped encoder-decoder backbone for ultrasound data. The Contrastive-TransUNet employs a custom contrastive loss tailored for video-style ultrasound sequences, promoting stable representation learning across temporally correlated frames. By combining these two models, our framework fuses generative and contrastive knowledge through a confidence-aware fusion head, in contrast to prior studies that typically rely on a single SSL paradigm.

(2) Label-efficient cross-domain knowledge transfer.
We demonstrate that a simple architecture pretrained with our framework achieves strong robustness across ultrasound datasets exhibiting substantial domain shift, all under the constraint of no data aggregation. Importantly, this holds even when the target dataset is completely unlabeled. This addresses an important real-world challenge: when only one dataset is labeled, our framework’s representations enable effective generalization to an unlabeled POCUS dataset collected with different probes or imaging characteristics.

\section{Methods}
\subsection{Dataset}
\subsubsection{Ultrasound Scanning}

The source and target domain ultrasound datasets were collected prospectively using different POCUS imaging systems with institutional ethics approval and informed patient and parental consent. Importantly, ultrasound acquisitions in both domains were conducted as additional examinations for research purposes rather than as part of the standard clinical diagnostic work-up. All patients underwent routine radiographic imaging as part of standard clinical care, which served as the diagnostic reference.

The source domain consists of a set of 21080 images from 91 pediatric patients, collected by a sonographer with 10 years of experience in 2021 at Stollery Children’s Hospital Emergency Department using a Philips Lumify L5–12 MHz probe. The target domain consists of a set of 22607 images from 136 pediatric patients, collected from a similar clinical population in the same geographic region but at a different time period (2023) by a different group of three medical students in 2023 at Stollery Children’s Hospital Emergency Department and Edmonton Cast Clinic using a TeleMED MicrUs Pro-L40S probe. All operators involved in both years received formal training in ultrasound scanning by medical professionals to ensure the ability to perform standardized examinations. Both devices have received FDA 510(k) clearance as diagnostic ultrasound systems.

Ultrasound examinations were performed on children aged 0-17 years who presented to the emergency department or cast clinic with wrist trauma.  In the source domain, the examination protocol included five cine sweeps in the following locations: (1) dorsal, including metaphysis, epiphysis, and first row of carpal bones; (2) proximal dorsal, including metaphysis and epiphysis; (3) radial, including metaphysis, epiphysis, and first row of carpal bones; (4) volar, including metaphysis, epiphysis, and first row of carpal bones; and (5) proximal volar, including metaphysis and epiphysis.  In the target domain, the acquisition protocol excluded the radial view due to its consistently lower image quality and limited additional diagnostic value, as most abnormalities observable in the radial view are also captured in the dorsal and volar views. Therefore, only four standard views (dorsal, proximal dorsal, volar, and proximal volar) were collected in the target domain. All patients in both the source and target domains had corresponding radiographs that served as the diagnostic reference standard and were used to ensure the accuracy of ultrasound interpretation and annotation. 

Only the TeleMED system provides access to raw ultrasound data, which is essential for developing our future AI-embedded mobile application. Consequently, while the source domain provides dense expert annotations, the TeleMED data serves as the primary target for system deployment.

\subsubsection{Image Analysis and Pixelwise Labeling}

For the source domain, bony region segmentation annotations were generated for each scan by a musculoskeletal sonographer with 10 years of experience under the supervision of an experienced radiologist using ITK-Snap. The manual annotation process was labor-intensive, requiring more than 30 minutes per video. To enable evaluation of the AI model in the target domain, sparse annotation was performed on roughly 12.5\% of the frames by a medical student under the supervision of an experienced radiologist using ITK-Snap. The specific use of these sparse annotations in model training and evaluation is described in the following \ref{subsec313} Dataset Splitting Details section.

Videos were converted into sequences of single frames, and only frames containing distinct and clear views, where the sonographer/medical student could confidently identify pathology, were kept, labeled and used in the study. Frames that did not capture the target region were excluded, as they could not support reliable expert annotation or meaningful diagnostic interpretation. Data were randomized and anonymized after collection. The two datasets were kept separate and treated as independent sites during model development and training. Further details of the datasets are provided in Table \ref{tab1}.

\begin{table}[t]
\centering
\caption{\normalsize Dataset details}
\label{tab1}
{\footnotesize
\begin{tabular}{l l l} 
\hline
Dataset & Source Domain & Target Domain\\
\hline
Year & 2021 & 2023 \\
\hline
POCUS Device & \makecell[l]{Philips Lumify\\L5--12 MHz} & \makecell[l]{TeleMED MicrUs\\Pro-L40S} \\
\hline
Centers & ER & ER, Cast Clinic \\
\hline
Sonographers & \makecell[l]{Group A: 10-year\\ experienced sonographer} & \makecell[l]{Group B: three \\medical students \\ with minimal training} \\
\hline
Labeling & \makecell[l]{A 10-year experienced \\sonographer}& A medical student \\
\hline
Annotation & All frames & 12.5\% of frames \\
\hline
Number of patients & 109 & 136 \\
\hline
Number of videos & 543 & 469 \\
\hline
Number of images & 24,902 & 22,607 \\
\hline
\makecell[l]{Number of frames\\ with annotation} & 24,902 & 2,843 \\
\hline
\end{tabular}}
\end{table}

\subsubsection{Dataset Splitting Details}
\label{subsec313}
For all experiments involving SSL pretraining, the datasets were split as follows:

The source domain was randomly divided into training (74 patients, 68\%), validation (17 patients, 16\%) and test (18 patients, 16\%) sets according to patient study IDs. All images in the source domain were accompanied by human annotations. The training set was used for downstream fine-tuning, the validation set was used to determine the best model, and the test set was used to evaluate model performance.

The target domain was divided into training (119 patients, 87.5\%) and test (17 patients, 12.5\%) sets based on patient study IDs. To maximize data utility, the training set leveraged the full breadth of available video frames containing bony regions for annotation-free pretraining. Conversely, the test set was restricted to the sparse subset of frames containing expert-labeled ground truth. Consequently, while the training set contains a substantially larger volume of images, the test set remains a highly curated cohort dedicated to inference. Evaluation was performed by comparing model predictions on these test frames against human annotations to assess segmentation performance.

Additional experimental setups included:

(1) Upper-bound evaluation: Train and validate on the target domain (sparse labels) and test on the target test set to see the best achievable performance when domain shift is removed.

(2) Lower-bound (Generalization) evaluation: Train and validate on the source domain (dense labels) and test on the target test set to measure cross-domain generalization without SSL pretraining.

Detailed splits are provided in Table \ref{tab2}.

Comprehensive details regarding the distribution of fracture-positive and fracture-negative cases across all datasets are provided in Table A of the Supplementary Materials. Furthermore, a detailed breakdown of the target domain test set, including the number of videos and selected frames per patient across different scanning views, is documented in Supplementary Materials Table B.

\begin{table}[t]
\centering
\begin{threeparttable}
\caption{\normalsize Dataset splitting details}
\label{tab2}
{\footnotesize
\begin{tabular}{p{1.8cm} p{1cm} p{1cm} p{1cm} p{1cm} p{1.5cm} p{1.5cm} p{1cm}} 
\hline
Dataset splitting & 
Src Train & 
Src Val & 
Src Test & 
Tgt Train & 
Tgt Train (Upper-bound)* & 
Tgt Val (Upper-bound)* & 
Tgt Test ** \\
\hline
\makecell[l]{Number \\of patients} & 74 & 17 & 18 & 119 & 102 & 17 & 17 \\
\hline
\makecell[l]{Number\\ of videos} & 370 & 84 & 89 & 407 & 346 & 61 & 62 \\
\hline
\makecell[l]{Number\\ of images} & 16,865 & 4,215 & 3,822 & 22,289 & 2,206 & 319 & 318 \\
\hline
\makecell[l]{Human\\ annotation} & Y & Y & Y & N & Y & Y & Y \\
\hline
\end{tabular}
}
\begin{tablenotes}
\footnotesize
\item Src: source domain; Tgt: target domain
\item[*] The dataset was split from column 5 target domain training set (Tgt Train).
\item[**] We only use test set images with human annotation for model evaluation.
\end{tablenotes}
\end{threeparttable}
\end{table}

All subsets include diagnostically challenging cases, such as healing and mild fractures, to better reflect real-world clinical scenarios.

\subsection{Model Architecture}
\begin{figure}[t]
    \centering  
    \includegraphics[width=\linewidth]{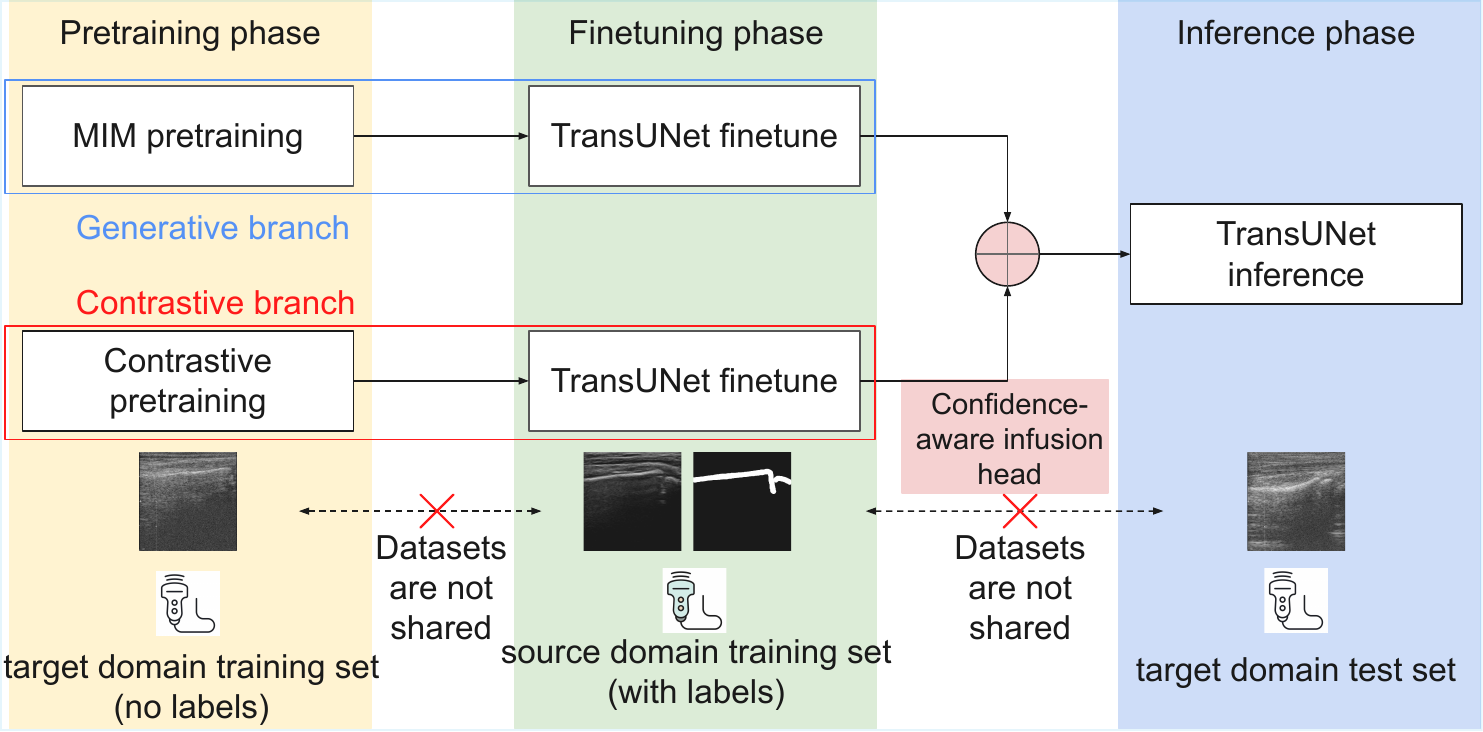}  
    \caption{Model architecture with two branches and a confidence-aware infusion head for final output. The model was self-supervised pretrained on the target domain via masked image modeling and contrastive learning, fine-tuned on the source domain with initialized weights, and finally fused through a confidence-aware infusion head for evaluation on the target-domain test set. Source domain and target domain datasets were kept strictly separate throughout the entire process.} 
    \label{fig2} 
\end{figure}

We adopted a two-parallel knowledge fusion framework for our model architecture, consisting of a generative branch and a contrastive branch, followed by an infusion strategy for final prediction, as shown in Figure \ref{fig2}.

\subsubsection{Generative Branch: MIM-TransUNet}
\label{subsec321}
\textit{MIM-TransUNet Pretraining}

For the generative branch, we designed an MIM-TransUNet architecture for pretraining on the target domain training set without any manual labels. This architecture was inspired by SimMIM \cite{xie2022simmim} where MIM is applied to images using a Vision Transformer for reconstruction. To better accommodate the personalized segmentation task, we replaced the SimMIM ViT reconstruction backbone with  TransUNet default architecture (R50-ViT-B16) \cite{chen2021transunet}, retaining its original structure to ensure compatibility with downstream segmentation. Specifically, the embedding was generated using a ResNet50 backbone, the encoder consisted of a ViT-B with 12 layers and 768-dimensional hidden states, and the decoder comprised four convolutional layers following the default TransUNet design. The segmentation head was implemented as a single convolutional layer, consistent with the default configuration. 

\begin{figure}[t]
    \centering  
    \includegraphics[width=\linewidth]{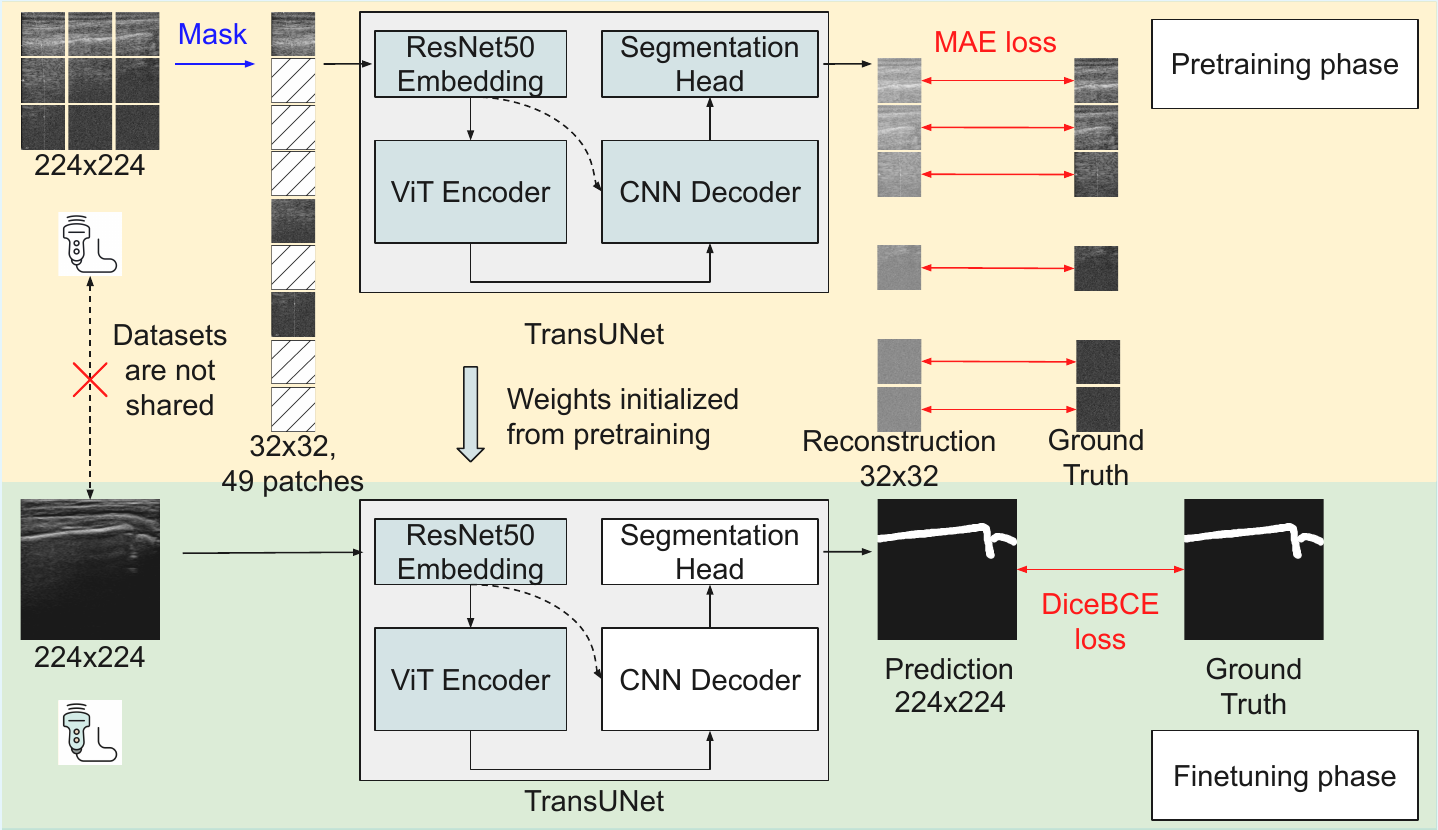}  
    \caption{Overview of the generative branch. Pretraining phase: Target domain training images were split into patches and randomly masked before being fed into the TransUNet model. The reconstruction of masked patches was compared with the ground-truth patches using MAE loss. Finetuning phase: The TransUNet encoder and embeddings were initialized with pretrained weights, and finetuned on the source domain training set.Source and target domain datasets were always kept separate.} 
    \label{fig3} 
\end{figure}

During pretraining, unlabeled images were padded to squares and resized to 224 $\times$ 224 pixels.  Following established practices in masked image modeling, 60\% of the patches were randomly masked with zeros, as this ratio has been shown to yield optimal representation learning in previous studies \cite{xie2022simmim, zhou2024simicl}. All patches, including unmasked patches and masked patches were used as input to the TransUNet network. The output of the network was the reconstruction of the original image, and the loss was computed only over the reconstructed regions of the masked patches using mean absolute error loss (MAE loss): $$\mathcal{L}_{\text{MAE}} = \frac{1}{|\mathcal{M}|} \sum_{p \in \mathcal{M}} |x_p - \hat{x}_p|,$$ where $\mathcal{M}$ denotes the set of masked patches, $x_p$ is the original patch, and $\hat{x}_p$ is the reconstructed patch. The generative branch architecture is shown in Figure \ref{fig3}.

The model was pretrained on a single NVIDIA L40S GPU with a batch size of 128. Optimization was performed using the AdamW optimizer with a learning rate of 0.0005 and a weight decay of 0.05 for 1200 epochs after hyperparameter tuning. The TransUNet was randomly initialized without using any pretrained weights. We observed consistent performance trends across different runs.

\textit{MIM-TransUNet Finetuning}

After pretraining, the TransUNet embedding and encoder were initialized with the pretrained weights, while the decoder was randomly initialized for downstream segmentation fine-tuning on the source domain training set, utilizing all available human annotation labels. We used DiceBCE loss for finetuning, which combines the Dice loss and binary cross-entropy (BCE) loss: $\mathcal{L}_{\text{Dice-BCE}} = \mathcal{L}_{\text{Dice}} + \mathcal{L}_{\text{BCE}}$, where the Dice loss is defined as$$\mathcal{L}_{\text{Dice}} = 1 - \frac{2 \sum_i \hat{y}_i y_i + \epsilon}{\sum_i \hat{y}_i + \sum_i y_i + \epsilon},$$
and the BCE loss is$$\mathcal{L}_{\text{BCE}} = -\frac{1}{N} \sum_i [y_i \log(\hat{y}_i) + (1 - y_i) \log(1 - \hat{y}_i)].$$ 
Here, $\hat{y}_i$ and $y_i$ represent the predicted probability and ground-truth label for pixel $i$, $N$ is the total number of pixels, and $ \epsilon$ is the constant 1 to avoid division by zero.

For downstream segmentation fine-tuning on the source domain, training was conducted on 4 NVIDIA L40S GPUs with a global batch size of 512, using stochastic gradient descent (SGD) with a learning rate of 0.0002 and weight decay of 0.05 for 200 epochs. The model checkpoint achieving the lowest DiceBCE loss on the source domain validation set was retained. All hyperparameters reported above were determined via hyperparameter tuning to optimize model performance.

\subsubsection{Contrastive Branch: Contrastive-TransUNet}
\label{subsec322}
\textit{Contrastive-TransUNet Pretraining}

For the contrastive branch Contrastive TransUNet, images were first augmented before being fed into the model. Each image was padded to a square and resized to 224 $\times$ 224 pixels, followed by random resized cropping with a scale range of 0.5 to 1.0 and an aspect ratio between 3/4 and 4/3 and rescaled to  224 $\times$ 224. Random horizontal flipping was applied, and random color jittering with brightness, contrast, and saturation set to 0.4 was applied with a probability of 0.8. The augmented images were then processed by the TransUNet R50 ViT B16 backbone during contrastive pretraining, with the encoder and embeddings preserved as described above. The encoder output features had a dimensionality of 768. These features were fed into our custom designed projection head. The projection head consisted of a linear layer maintaining the same input and output dimensions followed by a ReLU activation, and a second linear layer projecting the features to 128 dimensions for similarity comparison.

For vector similarity comparison, we proposed MT-NXent, a novel extension of the standard NT-Xent contrastive loss in SimCLR \cite{chen2020simple}, which incorporates temporal negative sample masking. During training, each batch was formed by randomly sampling images from the entire target domain training set, without restricting selection to frames from a single video. Let $\mathbf{z}_i$ and $\mathbf{z}'_i$ denote the normalized feature embeddings of an original image and its augmented variant, respectively. The embeddings are concatenated to form a combined set $\mathbf{Z} = [\mathbf{z}_1, \dots, \mathbf{z}_B, \mathbf{z}'_1, \dots, \mathbf{z}'_B]$. Pairwise cosine similarities are computed as $$s_{kl} = \frac{\mathbf{z}_k \cdot \mathbf{z}_l}{\tau}, \quad k,l = 1, \dots, 2B$$
where ${\tau}$ is a temperature parameter.

To avoid trivial correlations due to temporal proximity within the same video, a mask $M \in \{0, 1\}^{2B \times 2B}$ was constructed such that:
$$M_{kl} = \begin{cases} 1, & \text{if } k = l \text{ or if } \mathbf{z}_k \text{ and } \mathbf{z}_l \text{ belong to the same video and } \\& |\text{frame}_k - \text{frame}_l| < \Delta t, \text{ excluding positive pairs} \\ 0, & \text{otherwise} \end{cases}$$
where $\Delta t$ is a predefined minimum frame gap. Masked similarities are replaced with a large negative value to exclude them from loss computation:
$$\tilde{s}_{kl} = \begin{cases} s_{kl}, & M_{kl} = 0 \\ -\infty, & M_{kl} = 1 \end{cases}$$
Positive pairs are defined only between each original image and its augmented variant. Let $y_k$ denote the index of the positive pair for sample $k$. The final loss is formulated as the standard cross-entropy over the masked similarity matrix:
$$\mathcal{L} = \frac{1}{2B} \sum_{k=1}^{2B} -\log \frac{\exp(\tilde{s}_{k, y_k})}{\sum_{l=1}^{2B} \exp(\tilde{s}_{k, l})}$$
This formulation ensures that only the desired positive pairs contribute to similarity learning, while temporally adjacent frames within the same video are excluded from the negative sample set to prevent trivial correlations and improve representation quality.

\begin{figure}[t]
    \centering  
    \includegraphics[width=\linewidth]{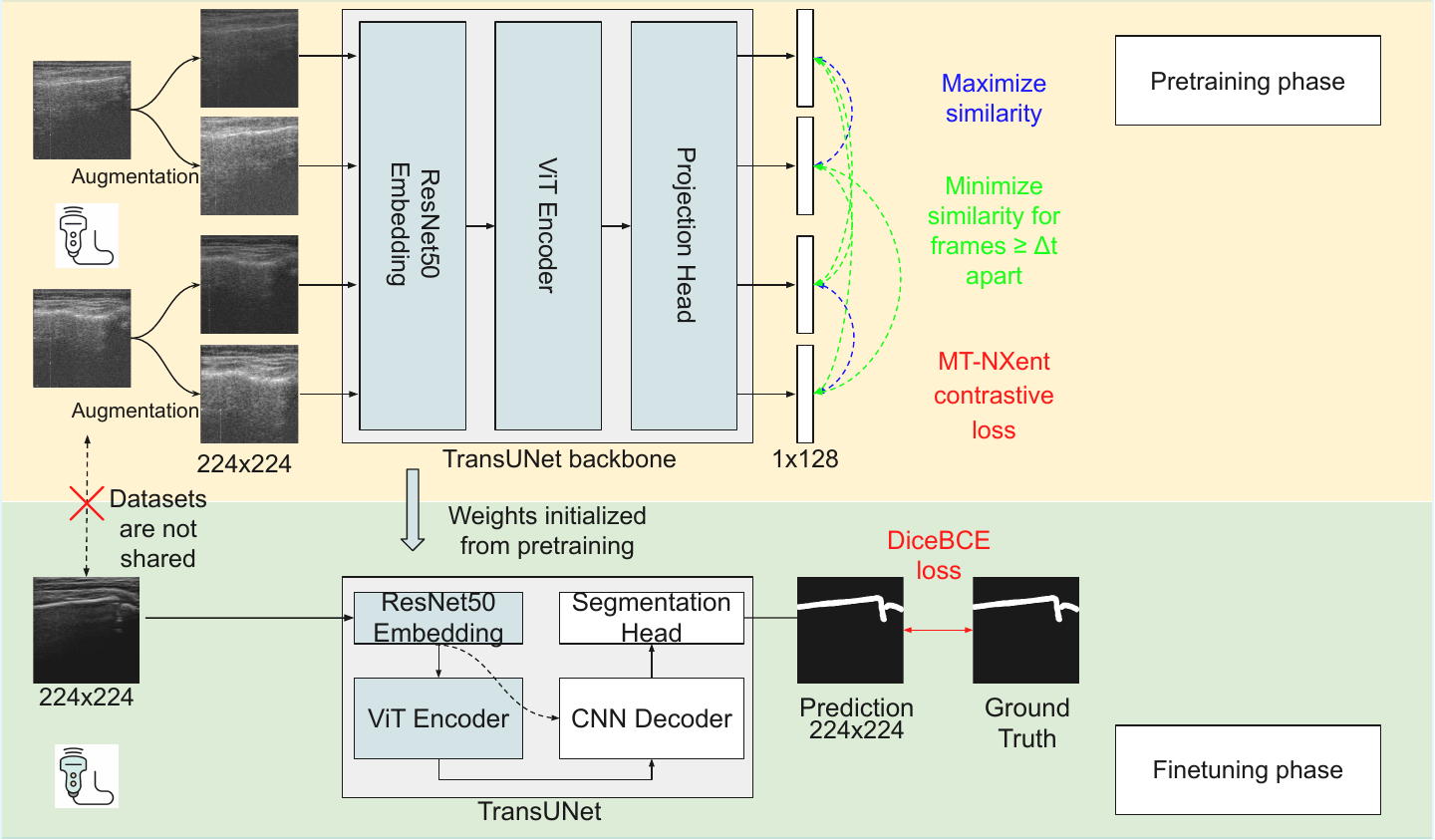}  
    \caption{Overview of the contrastive branch. Pretraining phase: Each target-domain training image was randomly augmented into two views before being fed into the TransUNet backbone. The encoder and embeddings were retained, followed by our simple projection head. The resulting 1×128 vectors were compared using our MT-NXent loss, which pulls vectors from the same original image closer and pushes vectors apart if they come from different images with a frame interval $\Delta $t. Finetuning phase: The TransUNet encoder and embeddings were initialized with pretrained weights and finetuned on the source-domain training set. Source and target domain datasets were always kept separate.} 
    \label{fig4} 
\end{figure}

Contrastive-TransUNet was pretrained on the target domain training set using MT-NXent loss to learn robust feature representations. The contrastive branch architecture is shown in Figure \ref{fig4}.

Pretraining was performed on 8 NVIDIA L40S GPUs. The effective global batch size was 1024, comprising 512 original images and their corresponding 512 augmented variants. This large batch size was employed because SimCLR relies on a substantial number of samples per iteration to effectively optimize the contrastive loss and achieve superior representation learning. The Adam optimizer was used with a learning rate of 0.00001 and a weight decay of 0.05 for 400 epochs after hyperparameter tuning. The TransUNet backbone was randomly initialized without pretrained weights. During contrastive training, the temperature parameter ${\tau}$ was set to 0.5 after hyperparameter tuning, and the predefined minimum frame gap $\ge \Delta$t was set to 0 to 30. 

\textit{Contrastive-TransUNet Finetuning}

The TransUNet model was initialized with the embeddings and encoder weights from the SSL pretraining, while the decoder and segmentation head were initialized randomly. We used the Dice-BCE loss to train our segmentation model. 

For downstream segmentation fine-tuning, 4 NVIDIA L40S GPUs were used with a global batch size of 512, SGD optimizer with a learning rate of 0.0002 and weight decay of 0.1, training for 200 epochs. The checkpoint achieving the lowest DiceBCE loss on the source domain validation set was retained. All hyperparameters reported above were determined via hyperparameter tuning to optimize model performance.

\subsubsection{Infusion Head}

After fine-tuning both the generative and contrastive branches independently, we performed inference on the target domain test set using both branches. We adopted a confidence-aware fusion strategy. For each pixel, the final output is computed as the average of both branches’ confidence-weighted predictions, $$\hat{y} = \frac{1}{2} (p_g \cdot c_g + p_c \cdot c_c)$$

where $p_g$ and $p_c$ denote the predicted probabilities from the generative and contrastive branches, respectively, and $c_g$ and $c_c$ represent their corresponding confidence scores. To ensure comparability between the confidence values from different branches, we further apply per-batch min–max normalization to $c_g$ and $c_c$, scaling them to the range [0, 1] prior to fusion.

We evaluated three confidence estimation strategies for each pixel.

(1) Softmax-probability–based Shannon entropy: $$c_i = 1 + \sum_k \hat{y}_{i,k} \log \hat{y}_{i,k}$$
where $\hat{y}_{i,k}$ is the predicted probability for class $k$ at pixel $i$. The entropy is reversed (1 + entropy) to obtain the confidence.

(2) Margin-based confidence: $$c_i = \hat{y}_{i,1} - \hat{y}_{i,2}$$
where $ \hat{y}_{i,1}$ and $ \hat{y}_{i,2}$ are the highest and second-highest predicted probabilities at pixel $i$. A larger margin indicates higher confidence.

(3) Simple average, where both branches are equally weighted without considering confidence.

For inference, all models applied a sigmoid activation to the output, followed by a fixed threshold of 0.5. Pixels with predicted probabilities $\geq$ 0.5 were classified as bony regions, while those below 0.5 were assigned to the background.

\subsection{Model Comparisons and Implementation Details}

We compared our approach with SimMIM and SimCLR, which are adaptations of the original SSL methods, both using the TransUNet backbone. These two methods share the same backbone structures as our MIM-TransUNet branch and Contrastive-TransUNet branch. The key difference is that SimCLR employs the original NT-Xent contrastive loss, whereas our method uses the redesigned MT-NXent contrastive loss. Implementation details are consistent with those described in Sections \ref{subsec321} and \ref{subsec322}.

For a fair comparison, all SSL methods follow the same training paradigm: pretraining on the target-domain training set, fine-tuning on the source-domain training set, and inference on the target-domain test set.

We additionally include two baselines:

Upper-bound baseline evaluation: TransUNet models were trained on the target domain training set, and the best-performing model on the target domain validation set with the lowest Dice BCE loss was saved. All available human annotations (sparse) were used for segmentation. Inference was performed on the target domain test set. This setup was designed to explore the upper limit of model performance when training, validation, and test data originate from the same source with no domain shift. 

Lower-bound(Generalization) baseline evaluation: TransUNet models were trained on the source domain training set, and the best model on the source domain validation set was saved, using all available human annotations (dense). Inference was then performed on the target domain test set to assess the generalization capability of the model without pretraining.

The two baseline methods described above were trained using four NVIDIA L40S GPUs with a global batch size of 512. Optimization was performed using SGD with a learning rate of 0.0002 and a weight decay of 0.05 after hyperparameter tuning, and the models were trained for 200 epochs.

\subsection{Evaluation Metrics}
We evaluated all models using the Dice Similarity Coefficient (DSC) and Intersection over Union (IoU). Higher values indicate better segmentation performance.

\section{Results}
\subsection{Model Performance}

We compared our method with state-of-the-art SSL models, SimMIM \cite{xie2022simmim} and SimCLR \cite{chen2020simple}. To ensure a fair and rigorous evaluation, all models were implemented using TransUNet as the common backbone. By maintaining a constant architectural setup, we isolate the experimental variables, ensuring that any observed performance gains are strictly attributable to the pre-training strategies rather than variations in model capacity or architecture. Results are shown in Table \ref{tab3}.

\begin{table*}[t]
\centering
\begin{threeparttable}
\caption{\normalsize Model performance comparison on the target domain test set}
\label{tab3}
{\footnotesize
\begin{tabular}{p{1.5cm} p{1.9cm} p{1.9cm} p{1.9cm} p{1.9cm} p{1.9cm}}
\hline
 & \multicolumn{3}{c}{Cross-Domain SSL} & \multicolumn{2}{c}{Conventional methods \cite{chen2021transunet}} \\
\hline
& Our model & SimMIM \cite{xie2022simmim} & SimCLR \cite{chen2020simple} & Lower-bound baseline (TransUNet) & Upper-bound baseline (TransUNet)  \\
\hline
Backbone & TransUNet & TransUNet & TransUNet & TransUNet & TransUNet \\
\hline
\makecell[l]{Pretrained\\ on\\ ImageNet} & N & N & N & Y & Y \\
\hline
\makecell[l]{Pretrained \\on target \\domain\\ dataset} & Y & Y & Y & N & N \\
\hline
Training data & The source domain training set & The source domain training set & The source domain training set & The source domain training set & The target domain training set \\
\hline
DSC & \textbf{0.7221} & 0.7116*** & 0.6992*** & 0.6605*** & 0.7798 \\
\hline
IoU & \textbf{0.5711} & 0.5580*** & 0.5425*** & 0.5063*** & 0.6391 \\
\hline
\end{tabular}
}
\begin{tablenotes}
\footnotesize
\item \textbf{Bold values} indicate the best performance among all methods except the upper-bound model trained without distribution shift. Asterisks denote statistically significant differences between our method and other methods (excluding the upper-bound model) based on the Wilcoxon signed-rank test (*** p < 0.001, ** p < 0.01, * p < 0.05).
\end{tablenotes}
\end{threeparttable}
\end{table*}

As presented in Table \ref{tab3}, the cross-domain SSL methods substantially outperform the non-SSL baseline with ImageNet-pretrained weights by 0.0616 in DSC and 0.0648 in IoU. Notably, our approach also demonstrates a significant performance margin over single SSL strategies, confirming the advantages of the proposed framework. The Wilcoxon signed-rank test confirms that these improvements over both the non-SSL baseline and single SSL methods are statistically significant, with p < 0.001. By exposing the model to target-domain imagery through our SSL framework, we effectively mitigate domain discrepancy and enhance feature robustness, leading to superior generalization under domain shift.

Nevertheless, the performance of all cross-domain SSL methods remains 0.0577 DSC and 0.0680 IoU lower than the upper-bound baseline, highlighting that training and testing on data from the same domain yields optimal performance due to the absence of domain shift. Among all compared SSL approaches, our method achieves the best performance, further validating the effectiveness of the proposed architecture.

\begin{figure}[t]
    \centering  
    \includegraphics[width=\linewidth]{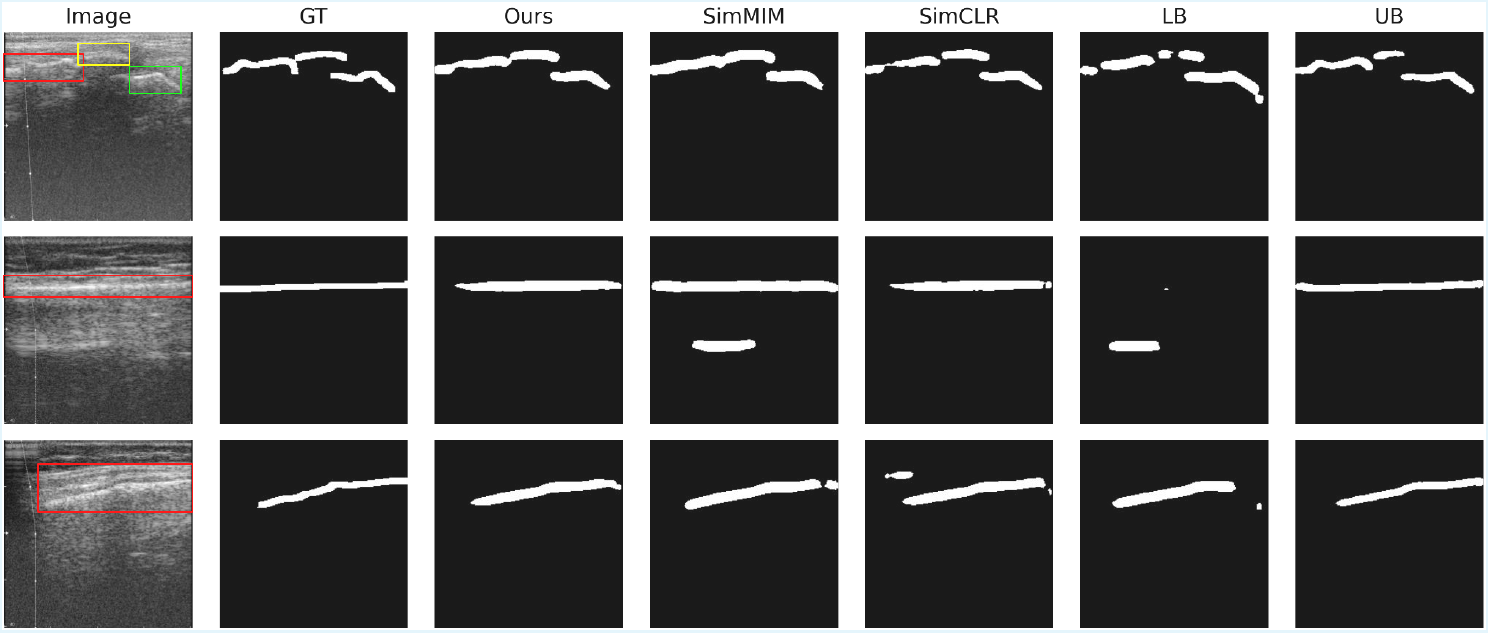}  
    \caption{Visualization of model predictions on the target-domain test set. GT: ground truth; LB: lower-bound baseline; UB: upper-bound baseline. In the leftmost image, red, yellow, and green rectangles mark the metaphysis, epiphysis, and carpal bones, respectively.} 
    \label{fig5} 
\end{figure}

The visualization of model predictions on the target-domain test set is shown in Figure \ref{fig5}. Compared to other SSL models, SimMIM and SimCLR, our model combines the accurate predictions from both while minimizing erroneous predictions. Lower-bound baseline, which was not pretrained on the target domain, performs suboptimally. In contrast, the upper-bound baseline, trained directly on the target domain without domain shift, achieves the best performance. Compared to the lower-bound baseline, these results demonstrate that our approach improves cross-domain robustness.

\begin{figure}[t]
    \centering  
    \includegraphics[width=\linewidth]{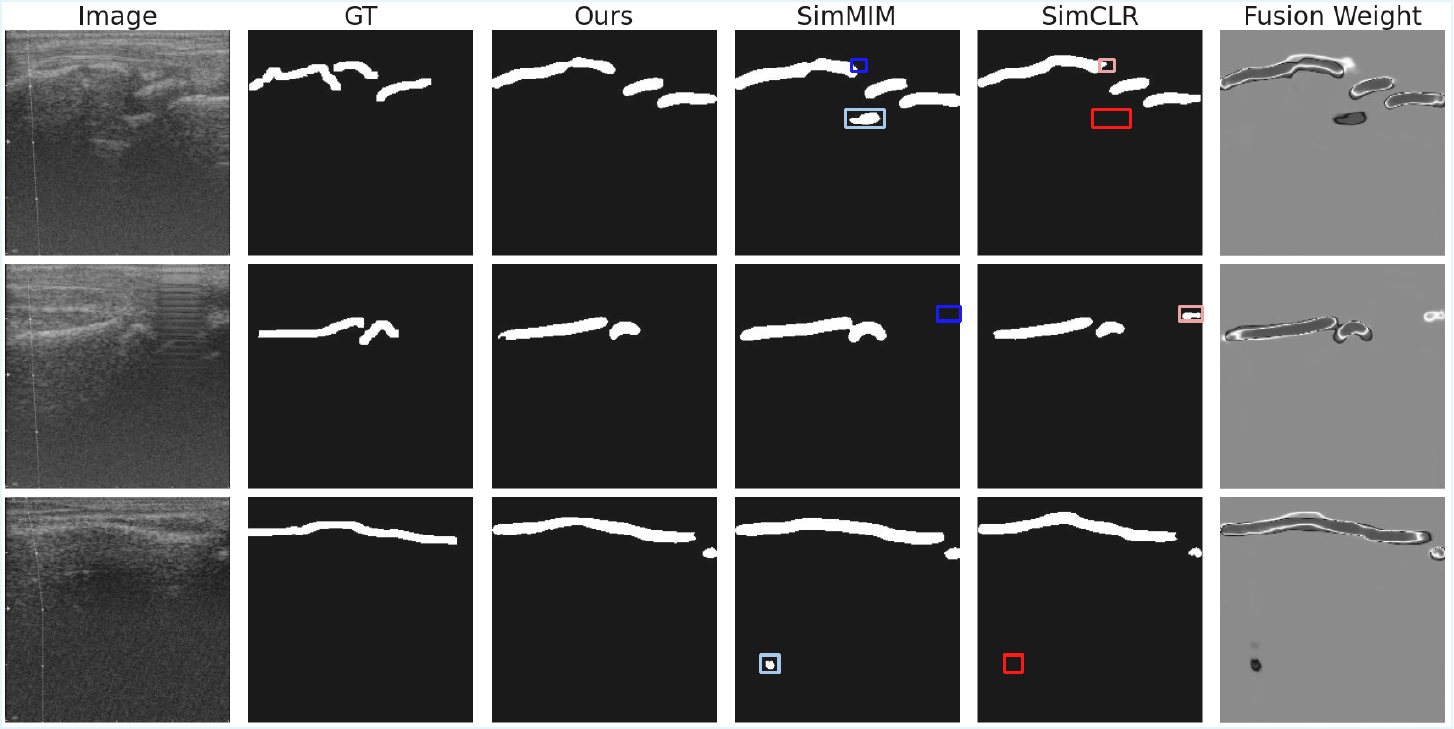}  
    \caption{ Visualization of predictions from the generative (SimMIM) and contrastive (SimCLR, with MT-NXent loss) branches and their confidence-aware fusion behavior. GT: ground truth. Dark blue and light red rectangles indicate areas where the generative branch has higher confidence and contributes more to the final prediction. Dark red and light blue rectangles indicate areas where the contrastive branch has higher confidence and contributes more to the final prediction. The fusion weight map highlights spatially adaptive branch contributions: lighter tones indicate higher reliance on the generative branch, while darker tones indicate stronger contribution from the contrastive branch.
} 
    \label{fig6} 
\end{figure}

To further validate the effectiveness of the confidence-aware infusion head, which dynamically combines predictions by estimating pixel-wise confidence scores from both the generative and contrastive branches, we visualize the fusion confidence maps of each branch. As shown in Figure \ref{fig6}, the “fusion weight” represents the relative contribution of the generative branch at each pixel, where lighter regions indicate greater reliance on the generative prediction and darker regions correspond to stronger contribution from the contrastive branch. For branch-specific predictions, blue and red annotations highlight regions with different dominance patterns. Specifically, blue and light red rectangles indicate areas where the generative branch is more dominant, while dark red and light blue rectangles indicate areas where the contrastive branch contributes more.

The visualization demonstrates that the ensemble does not systematically favor a single branch. Instead, it adaptively adjusts the contribution of each branch according to the pixel-wise confidence, resulting in a spatially dynamic fusion behavior. Consequently, the final prediction (“Ours”) is more consistent with the ground truth, benefiting from the complementary strengths of both branches.

Although our main goal is to build a robust model on the target domain dataset without human annotation, we also evaluated the model performance on the source domain test set to assess its robustness on data from the same domain as the training set. The results are presented in Table \ref{tab4}. Here, the “lower-bound baseline” and “upper-bound baseline” correspond to the minimum and maximum performance on the target domain test set, respectively.

Among all evaluated models, our model achieves the highest quantitative performance. Notably, our model demonstrates statistically significant improvements over all competing methods, with the exception of SimCLR, where performance was comparable. Moreover, all SSL models outperform the model trained solely on the source domain training set, indicating that incorporating additional data from the target domain improves the model’s robustness on the source domain. As expected, the upper-bound baseline, which is trained only on the target domain, exhibits the poorest performance on the source domain test set among the five models. Overall, the results in Tables \ref{tab3} and \ref{tab4} demonstrate that our model achieves superior performance across both source and target domains, aligning with the goal of obtaining a model that is strong and robust across datasets without sharing raw data during training.

\begin{table*}[t]
\centering
\begin{threeparttable}
\caption{\normalsize Model performance comparison on the source domain test set}
\label{tab4}
{\footnotesize
\begin{tabular}{p{1.5cm} p{1.9cm} p{1.9cm} p{1.9cm} p{1.9cm} p{1.9cm}}
\hline
 & \multicolumn{3}{c}{Cross-Domain SSL} & \multicolumn{2}{c}{Conventional methods \cite{chen2021transunet}} \\
\hline
& Our model & SimMIM \cite{xie2022simmim} & SimCLR \cite{chen2020simple}& Lower-bound baseline (TransUNet) & Upper-bound baseline (TransUNet) \\
\hline
Backbone & TransUNet & TransUNet & TransUNet & TransUNet & TransUNet \\
\hline
\makecell[l]{Pretrained\\ on\\ ImageNet} & N & N & N & Y & Y \\
\hline
\makecell[l]{Pretrained \\on the \\ target \\domain\\ dataset} & Y & Y & Y & N & N \\
\hline
Training data 
& The source domain training set 
& The source domain training set 
& The source domain training set 
& The source domain training set 
& The target domain training set \\
\hline
DSC 
& \textbf{0.8535} 
& 0.8519** 
& 0.8525 
& 0.8473*** 
& 0.7298*** \\
\hline
IoU 
& \textbf{0.7489} 
& 0.7471** 
& 0.7486 
& 0.7416*** 
& 0.5808*** \\
\hline
\end{tabular}
}
\begin{tablenotes}
\footnotesize
\item \textbf{Bold values} indicate the best performance for each metric. Asterisks denote the statistical significance of the difference between the current method and our model (the best-performing model) in bold based on the Wilcoxon signed-rank test  (*** p < 0.001, ** p < 0.01, * p < 0.05).
\end{tablenotes}
\end{threeparttable}
\end{table*}

Our method achieves an inference time of approximately 30 frames per second (FPS), corresponding to 33ms per frame, on an NVIDIA L40S GPU, with a computational cost of about 101.4 GFLOPs. Since inference is performed using only two TransUNet models in parallel, the overall design remains relatively lightweight and is suitable for deployment in cloud-based or industrial applications.

\subsection{Ablation Study}
We conducted ablation studies to investigate the contribution of each component in the proposed model. 

We investigated the effect of the temperature parameter ${\tau}$ in the contrastive branch. Following the original SimCLR framework, we evaluate four settings: 0.01, 0.1, 0.5, and 1. The predefined minimum frame gap $\Delta$t is set to 0. The model is first pretrained on the target-domain training set, then fine-tuned on the source-domain training set, and finally evaluated on the target-domain test set. All other hyperparameters are kept consistent with Section 3.2.2 Contrastive Branch: Contrastive-TransUNet after hyperparameter tuning. Table \ref{tab5} reports the downstream segmentation performance under different temperature settings.

\begin{table}[t]
\centering
\begin{threeparttable}
\caption{\normalsize Impact of the temperature parameter $\tau$ during pretraining on downstream performance on the target domain test set}
\label{tab5}
{\footnotesize
\begin{tabular}{p{2.2cm} p{1.6cm} p{1.6cm} p{1.6cm} p{1.6cm}}
\hline
$\tau$
& 0.01 
& 0.1 
& 0.5 
& 1 \\
\hline
DSC 
& 0.6958 
& 0.6840*** & \textbf{0.6992} 
& 0.6862*** \\
\hline
IoU 
& 0.5403 
& 0.5247*** & \textbf{0.5425} 
& 0.5272*** \\
\hline
\end{tabular}
}
\begin{tablenotes}
\footnotesize
\item \textbf{Bold values} indicate the best performance for each metric. Asterisks denote the statistical significance of the difference between the current method and the best-performing model in bold based on the Wilcoxon signed-rank test (*** $p < 0.001$, ** $p < 0.01$, * $p < 0.05$).
\end{tablenotes}
\end{threeparttable}
\end{table}

When ${\tau}$=0.5, the model achieves the best performance, followed closely by ${\tau}$=0.01, with no statistically significant difference between the two. In contrast, setting ${\tau}$=0.1 or ${\tau}$=1 leads to a clear degradation in performance compared to ${\tau}$=0.5. Therefore, we select ${\tau}$=0.5 as the default temperature in our experiments.

We evaluated the effectiveness of the proposed MT-NXent loss in comparison with the standard NT-Xent loss used in SimCLR, as well as the impact of the number of adjacent masked frames used during pretraining in our dataset. The results are summarized in Table \ref{tab6}.

\begin{table}[t]
\centering
\begin{threeparttable}
\caption{\normalsize Impact of the number of adjacent masked frames during pretraining on downstream performance on the target domain test set}
\label{tab6}
{\footnotesize
\begin{tabular}{p{2.2cm} p{1.4cm} p{1.4cm} p{1.4cm} p{1.4cm} p{1.4cm}}
\hline
Min frames 
& 0 (SimCLR) \cite{chen2020simple}
& 10 
& 15 
& 20 
& 30 \\
\hline
DSC 
& 0.6992*** 
& 0.7043*** 
& \textbf{0.7181} 
& 0.7072*** 
& 0.6955*** \\
\hline
IoU 
& 0.5425*** 
& 0.5485*** 
& \textbf{0.5663} 
& 0.5527*** 
& 0.5384*** \\
\hline
\end{tabular}
}
\begin{tablenotes}
\footnotesize
\item \textbf{Bold values} indicate the best performance for each metric. Asterisks denote the statistical significance of the difference between the current method and the best-performing model in bold based on the Wilcoxon signed-rank test (*** p < 0.001, ** p < 0.01, * p < 0.05).
\end{tablenotes}
\end{threeparttable}
\end{table}

As the number of masked adjacent frames increases, model performance on the target dataset improves from 0.6992 DSC to 0.7181 DSC, indicating more effective representation learning. However, performance plateaus at 15 masked frames and gradually decreases when a larger number of frames is masked. These results suggest that masking an appropriate number of adjacent frames during pretraining enables the model to learn target-domain representations more efficiently by better distinguishing between different images, thereby leading to improved downstream performance. The statistical validity of these improvements is further confirmed by the Wilcoxon signed-rank test, which yields significant results with p < 0.001 across these performance gains.

We also validated the effectiveness of the proposed confidence-aware infusion head. We evaluated a simple average ensemble, Shannon-entropy–based confidence-aware fusion, and margin-based confidence-aware fusion on the source domain validation set. The method achieving the best performance on the target domain validation set was then further tested on the target domain test set. The results are summarized in Table \ref{tab7}. Compared to a single branch, the ensemble yields a better performance. The Wilcoxon signed-rank test confirms that the improvements achieved by our confidence-aware fusion over both the single branches and the simple average baseline are statistically significant (p < 0.05 to p<0.001). These results indicate that ensemble strategies can improve performance over a single branch when an appropriate fusion method is employed.
\begin{table}[t]
\centering
\begin{threeparttable}
\caption{\normalsize The impact of the confidence-aware infusion head on the target domain test set}
\label{tab7}
{\footnotesize
\begin{tabular}{p{2.8cm} p{2.4cm} p{2.4cm} p{2.4cm}}
\hline
 & Our model 
 & SimMIM \cite{xie2022simmim}
 & \makecell[l]{SimCLR, with \\MT-NXent loss} \\
\hline
MIM branch 
 & Y & Y & N \\
\hline
Contrastive branch 
 & Y & N & Y \\
\hline
Ensemble method 
 & \makecell[l]{Softmax-\\probability-\\based\\ Shannon entropy} 
 & -- 
 & -- \\
\hline
DSC 
 & \textbf{0.7221} 
 & 0.7116*** 
 & 0.7181** \\
\hline
IoU 
 & \textbf{0.5711} 
 & 0.5580*** 
 & 0.5663* \\
\hline
\end{tabular}
}
\begin{tablenotes}
\footnotesize
\item \textbf{Bold values} indicate the best performance for each metric. Asterisks denote the statistical significance of the difference between the current method and the best-performing model in bold based on the Wilcoxon signed-rank test (*** p < 0.001, ** p < 0.01 * p < 0.05).
\end{tablenotes}
\end{threeparttable}
\end{table}

To investigate whether performance gains stem from using two distinct branches, we conducted a homogeneous dual-branch setting, using  Shannon-entropy–based confidence-aware fusion. Specifically, both the generative and contrastive branches adopt the same type of branch, each configured with its best-performing configuration. Each model was trained multiple times, and the top two runs with the best performance were selected. For the contrastive branch, we further examined the effect of temporal frame gap consistency by comparing two settings: identical frame gaps (15 frames for both branches) versus mixed frame gaps (15 and 20 frames). In both cases, we report the top two best-performing runs.

The results in Table \ref{tab8} show that, numerically, individual homogeneous branches can achieve slightly higher performance than the heterogeneous setup (best homogeneous DSC: 0.7181 and 0.7163; heterogeneous DSC: 0.7181 and 0.7116). However, after ensembling, the homogeneous setting still underperforms the heterogeneous one, with the difference being statistically significant. This indicates that combining different types of branches is more effective than using two identical branches, leading to greater overall performance gains.

\begin{table}[t]
\centering
\begin{threeparttable}
\caption{\normalsize The impact of heterogeneous branch ensemble on the target domain test set}
\label{tab8}
\setlength{\tabcolsep}{1.2pt} 
{\scriptsize
\begin{tabular*}{\textwidth}{@{\extracolsep{\fill}} l ccc ccc ccc c}
\hline
 & \multicolumn{3}{c}{\begin{tabular}[c]{@{}c@{}}Homogeneous generative\\ branch: SimMIM\end{tabular}} 
 & \multicolumn{3}{c}{\begin{tabular}[c]{@{}c@{}}Homogeneous contrastive\\ branch: SimCLR, \\with MT-NXent loss\end{tabular}} 
 & \multicolumn{3}{c}{\begin{tabular}[c]{@{}c@{}}Homogeneous contrastive\\ branch: SimCLR, \\with MT-NXent loss\end{tabular}} 
 & \begin{tabular}[c]{@{}c@{}}Hetero\\ dual \\branch \end{tabular} \\
\cline{2-4} \cline{5-7} \cline{8-10} \cline{11-11}
 & \begin{tabular}[c]{@{}c@{}}Best\\ model 1\end{tabular} 
 & \begin{tabular}[c]{@{}c@{}}Best\\ model 2\end{tabular} 
 & Ens.
 & \begin{tabular}[c]{@{}c@{}}Best\\ model 1,\\ 15 frames\\ gap\end{tabular} 
 & \begin{tabular}[c]{@{}c@{}}Best\\ model 2,\\ 15 frames\\ gap\end{tabular} 
 & Ens.
 & \begin{tabular}[c]{@{}c@{}}Best\\ model 1,\\ 15 frames\\ gap\end{tabular} 
 & \begin{tabular}[c]{@{}c@{}}Best\\ model 2,\\ 20 frames\\ gap\end{tabular} 
 & Ens.
 & Ens. \\
\hline
DSC 
& \begin{tabular}[c]{@{}c@{}}0.7116\\ *** \end{tabular} & \begin{tabular}[c]{@{}c@{}}0.7106\\ *** \end{tabular} & \begin{tabular}[c]{@{}c@{}}0.7193\\ *** \end{tabular} 
& \begin{tabular}[c]{@{}c@{}}0.7181\\ ** \end{tabular} & \begin{tabular}[c]{@{}c@{}}0.7163\\ *** \end{tabular} & \begin{tabular}[c]{@{}c@{}}0.7200\\ * \end{tabular} 
& \begin{tabular}[c]{@{}c@{}}0.7181\\ ** \end{tabular} & \begin{tabular}[c]{@{}c@{}}0.7072\\ *** \end{tabular} & \begin{tabular}[c]{@{}c@{}}0.7191\\ ** \end{tabular} 
& \textbf{0.7221} \\
\hline
IoU 
& \begin{tabular}[c]{@{}c@{}}0.5580\\ *** \end{tabular} & \begin{tabular}[c]{@{}c@{}}0.5567\\ *** \end{tabular} & \begin{tabular}[c]{@{}c@{}}0.5665\\ *** \end{tabular} 
& \begin{tabular}[c]{@{}c@{}}0.5663\\ * \end{tabular} & \begin{tabular}[c]{@{}c@{}}0.5629\\ *** \end{tabular} & \begin{tabular}[c]{@{}c@{}}0.5680\\ * \end{tabular} 
& \begin{tabular}[c]{@{}c@{}}0.5663\\ * \end{tabular} & \begin{tabular}[c]{@{}c@{}}0.5527\\ *** \end{tabular} &\begin{tabular}[c]{@{}c@{}}0.5670\\ ** \end{tabular}
& \textbf{0.5711} \\
\hline
\end{tabular*}
}
\begin{tablenotes}
\footnotesize
\item Ens.: Ensemble
\item \textbf{Bold values} indicate the best performance for each metric. Asterisks denote the statistical significance of the difference between the current method and the best-performing model in bold based on the Wilcoxon signed-rank test (*** $p < 0.001$, ** $p < 0.01$, * $p < 0.05$).
\end{tablenotes}
\end{threeparttable}
\end{table}

\section{Discussion}
Motivated by challenges generalizing ultrasound models across hardware, users and sites, we built a novel SSL framework that combines MIM and contrastive learning for cross-domain segmentation of wrist ultrasound datasets. Instead of conventional strategies for handling domain shift, we adopt a cross-domain SSL strategy: it uses unlabeled target-domain data for initial training, and is then fine-tuned on labeled source-domain data, before finally being applied back to the target domain for testing. This pretraining phase enables the model to learn robust, domain-specific structural representations, thereby enhancing its generalization capabilities and ensuring stable performance when transitioning from source-domain supervision to target-domain inference.

Due to differences in hardware design and signal-processing pipelines between the two probes, the resulting images display distinct visual characteristics, leading to a clear domain shift between the datasets. Besides, the source domain data were collected by a single operator, whereas the target domain involved three operators with varying levels of skill and experience, further contributing to the domain shift.

A possible explanation for the observed improvement in source domain performance after pretraining on the target domain dataset is that, although the two datasets differ considerably due to probe signal processing and other imaging techniques, they share intrinsic similarities: both consist of wrist ultrasound images. Consequently, pretraining on the target domain enables the model to better capture the anatomical structures of the wrist, including bone shapes and relevant image features. Compared to using pretrained weights from ImageNet, SSL-based pretraining on wrist ultrasound images allows the model to develop a better understanding, thereby enhancing its performance on the source domain dataset.

The selection of an appropriate number of adjacent frames to mask during contrastive loss computation plays a critical role in model performance. If adjacent frames are too similar and not masked, the model may become confused and fail to learn discriminative features. Conversely, masking too many frames reduces the number of effective training samples, which can hinder feature learning.

Ensemble strategies can further enhance performance compared to using a single branch. Notably, the Softmax-probability–based Shannon entropy confidence-aware fusion is both convenient and computationally efficient, as it derives pixel-wise confidence directly from the model’s output probabilities without requiring additional computation.

Beyond performance improvements, the proposed framework is particularly well suited for federated learning scenarios commonly encountered in medical imaging. In many real-world settings, ultrasound data are stored locally at each clinical center and cannot be shared due to privacy, regulatory, or institutional constraints. Our two-step design naturally aligns with the federated learning paradigm: each site first performs self-supervised pretraining locally using its own unlabeled data, followed by fine-tuning on locally available annotated data, while sharing only model weights throughout training and inference without exposing raw data. This eliminates the need for data aggregation while enabling collaborative learning across institutions. This design makes the framework highly practical for large-scale, multi-center deployment, where heterogeneous data distributions and strict data-sharing restrictions are the norm.

Despite these improvements, our study has several limitations. Further investigation is needed to explore whether incorporating additional SSL strategies could further enhance the model’s robustness and performance.  In addition, our method has only been validated on wrist ultrasound datasets. Future work should assess its generalizability to other datasets, including not only domain shifts caused by different probes, but also anatomical variations (e.g., wrist versus elbow) to determine the extent to which the proposed approach can maintain or enhance performance across diverse scenarios.

The proposed MT-NXent loss relies on a fixed temporal frame gap to construct negative samples, which may not fully account for variations in scanning speed in real clinical practice. In our study, the dataset was collected by uniformly trained operators following a standardized acquisition protocol, thereby reducing variability in scanning speed. As a result, the temporal gap serves as a reasonable proxy for spatial displacement in our setting. Nevertheless, this assumption may not generalize to more diverse clinical scenarios, and the use of a fixed temporal gap remains a limitation of our approach. Future work will investigate more adaptive strategies, such as spatial- or content-aware sampling, to improve robustness under varying acquisition conditions.

Importantly, we emphasize that our method is not intended to replace established clinical standards of care, such as radiography and expert radiological assessment, but rather to serve as a complementary tool that may assist in emergency department triage or preliminary screening in resource-limited or remote settings. In particular, we acknowledge that false negative cases, meaning missed fracture detections, may have significant clinical consequences, especially for severe fractures that require timely intervention and follow-up treatment. Therefore, such failure modes should be carefully considered and mitigated before clinical deployment. While prior studies have reported promising results for AI-based ultrasound analysis in wrist fracture detection, these approaches should be interpreted with caution in clinical contexts, as diagnostic decisions and treatment planning in standard practice require comprehensive evaluation by certified clinicians and established imaging modalities. In this regard, we view the potential clinical contribution of our approach as supporting point-of-care decision-making and improving accessibility in settings where radiography may be delayed or unavailable, rather than replacing standard diagnostic pathways.

\section{Conclusion}

We propose a novel cross-domain SSL framework that combines MIM and contrastive learning for cross-domain segmentation of wrist ultrasound datasets. In this framework, we develop the MIM-TransUNet and redesign the contrastive loss for SimCLR, named MT-NXent, specifically for video-frame pretraining. Also adding a confidence-aware infusion head ensemble, our approach integrates multiple components to achieve performance exceeding state-of-the-art cross-domain SSL models.

Our method addresses the challenging scenario of unlabeled target domains, where the model is trained on labeled source-domain data yet achieves robust segmentation performance on the target domain. The proposed framework does not require data aggregation, making it particularly suitable for settings with strict data governance and privacy constraints. This provides a practical solution to the common problem of limited annotated medical data and enables effective model training across different ultrasound probes and clinical settings. The proposed framework demonstrates encouraging potential for specific applications in medical imaging, particularly in scenarios where annotations are scarce and cross-domain robustness is a primary concern.

\section*{Acknowledgements}
Dr. Jacob L. Jaremko is supported by the Canada CIFAR AI Chair, and his academic time is made available by Medical Imaging Consultants (MIC), Edmonton, Canada. We acknowledge the funding support of Alberta Innovates AICE Concepts (US-AID), Alberta Innovates Graduate Student Scholarships, WCHRI Graduate Studentship, IC-IMPACTS, TD Bank, AMII, CIFAR, and the Alberta Emergency Strategic Clinical Network. We thank the Digital Research Alliance of Canada for providing us with computational resources including high-power Graphical Processing Units (GPU) that were used for training and testing our deep learning models.

\bibliographystyle{elsarticle-num} 
\bibliography{refs} 
\end{document}